\documentclass[letterpaper, 10pt, conference]{ieeeconf}  

\IEEEoverridecommandlockouts                              

\usepackage{times}  
\usepackage{helvet}  
\usepackage{courier}  
\usepackage[hyphens]{url}  
\usepackage{graphicx} 
\usepackage{caption} 
\usepackage{amsmath}
\usepackage{amsfonts}
\usepackage{algorithm}
\usepackage{algorithmic}
\usepackage{subcaption}
\usepackage{ctable}
\newtheorem{lemma}{Lemma}
\newcommand{\cmmnt}[1]{\ignorespaces}

\usepackage{newfloat}
\usepackage{listings}
\title{\LARGE \bf
$CoDe:$ A Cooperative and Decentralized Collision Avoidance Algorithm for Small-Scale UAV Swarms Considering Energy Efficiency
}

\author{Shuangyao Huang$^{1}$, Haibo Zhang$^{2}$ and Zhiyi Huang$^{2}$
\thanks{$^{1}$Shuangyao Huang is with the School of Computing, 133 Union Street East, University of Otago, Dunedin 9016, New Zealand, and also with the School of Internet of Things, Xi'an Jiaotong-Liverpool University, Suzhou 215400, China {\tt\small shuangyao.huang@xjtlu.edu.cn}}%
\thanks{$^{2}$Haibo Zhang and Zhiyi Huang are with the School of Computing, 133 Union Street East, University of Otago, Dunedin 9016, New Zealand {\tt\small \{haibo.zhang, zhiyi.huang\}@otago.ac.nz}}%
\thanks{This work is jointly supported by XJTLU RDF-23-02-026, RDF-22-02-047, and RDF-22-02-099. }%
}

\begin{document}

\maketitle
\thispagestyle{empty}
\pagestyle{empty}

\begin{abstract}

	This paper introduces a cooperative and decentralized collision avoidance algorithm ($CoDe$) for small-scale UAV swarms consisting of up to three UAVs. $CoDe$ improves energy efficiency of UAVs by achieving effective cooperation among UAVs. Moreover, $CoDe$ is specifically tailored for UAV's operations by addressing the challenges faced by existing schemes, such as ineffectiveness in selecting actions from continuous action spaces and high computational complexity. $CoDe$ is based on Multi-Agent Reinforcement Learning (MARL), and finds cooperative policies by incorporating a novel credit assignment scheme. The novel credit assignment scheme estimates the contribution of an individual by subtracting a baseline from the joint action value for the swarm. The credit assignment scheme in $CoDe$ outperforms other benchmarks as the baseline takes into account not only the importance of a UAV's action but also the interrelation between UAVs. Furthermore, extensive experiments are conducted against existing MARL-based and conventional heuristic-based algorithms to demonstrate the advantages of the proposed algorithm. 

\end{abstract}

\section{INTRODUCTION}

Multi-rotor Unmanned Aerial Vehicles (UAV) are studied in this paper owning to their prevalence in research and industry. A UAV swarm is a fleet of UAVs that work together in a common task. Small-scale swarms consisting of up to three UAVs are sufficient for most tasks such as search and rescue {\cite{application4}}, tracking and monitoring {\cite{application4-1}}, \cmmnt{data collection {\cite{application4-4}},} and delivery {\cite{application4-4}}. However, the lack of energy-efficient collision avoidance algorithms for small-sale UAV swarms is preventing these applications from blossoming. Energy efficiency is critical for UAV swarm applications as the flight time of off-the-shelf UAVs \cmmnt{\cite{MATRICE600}} is limited by onboard power supply. Energy-efficient collision avoidance algorithms for UAV swarms is challenging in achieving cooperation, selecting actions from continuous action spaces and reducing computational complexity. 

Conventional methods for UAV collision avoidance typically suffer from difficulties in addressing short-sightedness and achieving cooperation. Geometry-based \cmmnt{\cite{vo, cone}}\cite{reciprocalvo}, virtual force-based \cmmnt{\cite{b11, b12}}\cite{b11}, and metaheuristic-based methods \cmmnt{\cite{pso1, pso2, pso3}}\cite{pso1} adopt a receding time horizon planning approach, which results in the low energy efficiency of UAVs due to zigzag trajectories caused by the short-sightedness of UAVs. 
Moreover, these methods are difficult to achieve cooperation among UAVs as they regard other UAVs as part of the environment when planning for each UAV. In metaheuristic-based methods, on the other hand, the search space's dimensionality becomes overwhelmingly complex when coordinating the planning for multiple UAVs simultaneously. The absence of cooperation further increases UAVs' energy consumption, as additional movements and maneuvers are needed to avoid collisions.  
In contrary to conventional methods, intelligent methods such as MARL-based approaches \cmmnt{\cite{iql_oda2, MADDPG_oda1, MADDPG_oda2, shapley_oda1}}\cite{MADDPG_oda1} train UAVs to maximize a long-term accumulative team reward, which addresses the short-sightedness of UAVs in receding time horizon planning approach. The cooperation of UAVs are achieved by maximizing a team reward through credit assignment schemes which estimate the contribution of each UAV in the swarm. 
However, existing credit assignment schemes are not suitable for UAV collision avoidance due to ineffectiveness in continuous action spaces and high computational complexity. 
For instance, policy-based MARL algorithms such as Counterfactual Multi-Agent Policy Gradients (COMA) \cite{COMA} and Shapley Q learning \cmmnt{\cite{Shapley, Shapley1}}\cite{Shapley} derive a counterfactual baseline to estimate the team reward assuming without the participation of each UAV. However, COMA is best suited for discrete action space, hence is not suitable for UAV operation which is performed by continuous actions such as throttle and motion angles. On the other hand, Shapley Q learning suffers from high computational complexity in deriving baselines, which makes it difficult for the UAVs to train cooperative policies. Moreover, their baselines either only consider the importance of UAVs' actions or the interrelation between UAVs, which may not accurately estimate the contribution of each UAV. On the contrary, value-based MARL algorithms such as Value Decomposition Network (VDN) \cite{VDN} and QMix \cite{Qmix} achieve credit assignment by decomposing the joint action value to individual values additively and linearly, respectively. The joint action values learned this way will result in unbounded divergence and lead to unexpected actions as proved in work \cite{divergence}. 
Other methods such as QTRAN \cite{Qtran} and Qplex \cite{Qplex} improve the stability of value decomposition networks with a richer family of network structures such as multi-head attention mechanisms. Despite this improvement, they maintain dependencies on assumptions regarding the relationships between joint value functions and individual value functions. 

This paper proposes a novel MARL-based \textit{Co}operative and \textit{De}centralized Collision Avoidance ($CoDe$) algorithm for small-sale UAV swarms. The objective of $CoDe$ is to improve energy efficiency of UAVs in collision avoidance by achieving effective cooperation among UAVs using policy-based MARL. The key contribution of $CoDe$ lies in the new credit assignment scheme that does not rely on any assumptions of value functions and applies to continuous action space, and its novel baseline that considers both the importance of a UAV's action and the interrelation between UAVs to best estimate the contribution of each UAV. 


For the rest part of this paper, related works are discussed in Section \ref{Relat}, followed by background in Section \ref{Relat}. After background knowledge, the application scenario and system model are introduced in Sections \ref{Appli} and \ref{System}, respectively. Most importantly, the algorithm of $CoDe$ is introduced in Section \ref{Algo} in detail, followed by evaluations in Section \ref{Exper}. Lastly, conclusions are drawn in Section \ref{Concl}. 

\section{Related Work} \label{Relat}
\subsection{Conventional Methods} 
Velocity Obstacle (VO) \cite{reciprocalvo} avoids collisions by selecting velocities outside a velocity pool consisting of velocities that would cause collisions with other UAVs or obstacles. While straightforward, VO relies on the assumption that obstacles can be modeled as a circle with safety distance being the radius. Therefore, VO does not work in complex environments where obstacles have particular shapes and cannot be molded as circles. 
Artificial Potential Fields (APF)-based approaches \cite{b11} provide an effective way of modeling complex environments as a two-dimensional differentiable potential field regardless of the numbers and shapes of obstacles. After the modeling, the UAVs are guided to avoid collisions by leveraging the gradients on the potential field. Nevertheless, the presence of local optima on the potential field, characterized by zero gradients, can result in UAVs becoming trapped inside, leading to energy-inefficient zig-zag trajectories. 
Metaheuristic-based approaches, such as swarm intelligence \cite{hyb5}, genetic algorithms \cite{geneticapplication1}, and graph path-finding algorithms \cite{pathfinding1, pathfinding2, pathfinding3}, address energy efficiency in collision avoidance by minimizing a cost function that considers energy consumption. Nonetheless, these methods face limitations imposed by the curse of dimensionality, which poses a challenge to the collaboration among swarm members, particularly when the dimensionality of the search space expands with the size of the swarm. 

\subsection{Intelligent Methods} 
State-of-the-art intelligent methods exploit Multi-Agent Reinforcement Learning (MARL) techniques to find cooperative collision avoidance strategies for UAVs, by modeling the sequential decision-making in collision avoidance as a Decentralized Partially Observable Markov Decision Process (Dec-POMDP). 
Some MARL-based approaches investigate Independent Q Learning (IQL) \cite{iql_oda1} and Multi-Agent Deep Deterministic Policy Gradient (MADDPG) \cite{MADDPG_oda1} to find cooperative policies for collision avoidance. Notably, these approaches rely on observation sharing but tend to yield sub-optimal policies due to the absence of effective credit assignment mechanisms. 
In contrast, other works investigate VDN \cite{vdn_oda1} and Qmix \cite{qmix_oda1} in cooperative collision avoidance. These methods embrace credit assignment strategies, which have a superior performance to observation-sharing approaches. Nevertheless, VDN and Qmix are value-based algorithms which rely on assumptions of value functions and fail to fully exploit global information during training, potentially resulting in sub-optimal policies. Moreover, the value decomposition in VDN and Qmix may result in unbounded divergence and may lead to unexpected behaviors of UAVs \cite{divergence}. 
Other than value-based MARL algorithms, policy-based algorithms such as Shapley Q-Learning \cite{shapley_oda1} and COMA \cite{coma_oda1} are also investigated in cooperative collision avoidance. These methods are assumption-free and make full advantage of global information in training. However, Shapley Q-Learning entails high computational complexity in credit assignment and COMA is best suited for discrete action space, making them unsuitable for UAV control scenarios that involve high-dimensional continuous action space. 

\section{Background} \label{Backg}

\subsection{Decentralized Partially Observable Markov Decision Process (Dec-POMDP)} 
A Dec-POMDP is defined by a tuple $(\mathcal{S}, \mathcal{A}, r, \mathcal{O}, \mathcal{P})$, where $\mathcal{S}$, $\mathcal{O}$ and $\mathcal{A}$ are state, observation and action space of the environment, respectively. $r$ is a reward signal and $\mathcal{P}$ is the state transition probability of the environment. 

At each step, each agent makes an observation $o\in \mathcal{O}$ on the environment, and selects an action $a\in \mathcal{A}$ based on its observation. Additionally, the environment also has a true state $s \in \mathcal{S}$.  
When all the agents execute their actions simultaneously, the environment transits from state $s$ to $s'$ with probability $P(s'|s, \boldsymbol{a}): \mathcal{S}\times\mathcal{A}\times\mathcal{S}\rightarrow[0,1]$, and returns a numerical reward $r(s, a): \mathcal{S}\times\mathcal{A}\rightarrow\mathbb{R}$, where $\boldsymbol{a}\equiv\{a^1, \cdots, a^{n}\}$ denotes the joint action of $n$ agents. In cooperative missions, all participating agents within the team receive the same reward, known as the team reward. 
Eventually, each agent maximizes a long-term team return defined by the accumulation of the team reward over time: $R_t = \sum_{l=0}^{\infty}\gamma^{l-1}r_{t+l}$, where $\gamma\in[0, 1)$ is a discount factor. 

\subsection{Multi-Agent Reinforcement Learning (MARL)} 

The overall MARL framework consists of two phases: off-line training and on-line execution. 
During off-line training, a centralized critic denoted as $f_{\omega}(\cdot)$ is developed to approximate the joint value $Q_{tot}$ for the entire swarm. The inputs to $f_{\omega}(\cdot)$ encompass the environmental state $s_t$, the joint observation $\boldsymbol{o}_t$, and the joint action $\boldsymbol{a}_t$ of all agents. 
The parameters of $f_{\omega}(\cdot)$ are updated by minimizing a temporal difference loss as in standard Q learning: 
\begin{equation} \label{criticloss} 
\begin{gathered} 
\mathcal{L}_{TD} = \mathbb{E}_{\pi_{\theta}}\left[
(y_t - f_{\omega}(t))^2\right], \\ 
y_t = r_t + \gamma \cdot f_{\bar{\omega}}(t+1), 
\end{gathered} 
\end{equation} 
where $\bar{\omega}$ are parameters of a target network periodically updated by $\omega$. The target critic has the same structure as the centralized critic but has its own parameters. The target critic is used to cut bootstrap in calculating temporal difference loss to stabilize the training process. 

On the other hand, the actor network $\pi_{\theta}$ is trained by cooperative policy gradient $g$: 
\begin{equation} \label{policygradient} 
\begin{gathered} 
g=\mathbb{E}_{\pi_{\theta}} \left[\nabla_{\theta}A_{\omega}(\pi_{\theta}(\tau_i))
\right], \\
\end{gathered} 
\end{equation}
where $\theta$ are parameters of actor networks, $A_{\omega}$ is an agent-specific advantage function and $\tau_i$ denotes observation history of agent $i$. The advantage function of agent $i$ measures its contribution in acquiring the joint value. The advantage function is defined by counterfactual baselines in policy-based approaches, and equals individual value in value-based approaches. 
Moreover, actor networks use observation history in input as they commonly use Recurrent Neural Network (RNN) to make up for the partial observability of agents caused by limited sensing ability. 

In contrary, it only requires the actor networks to generate actions in the execution phase. 


\section{Application Scenario} \label{Appli}

\begin{figure}[t]
	\centering
	\includegraphics[width=0.4\textwidth]{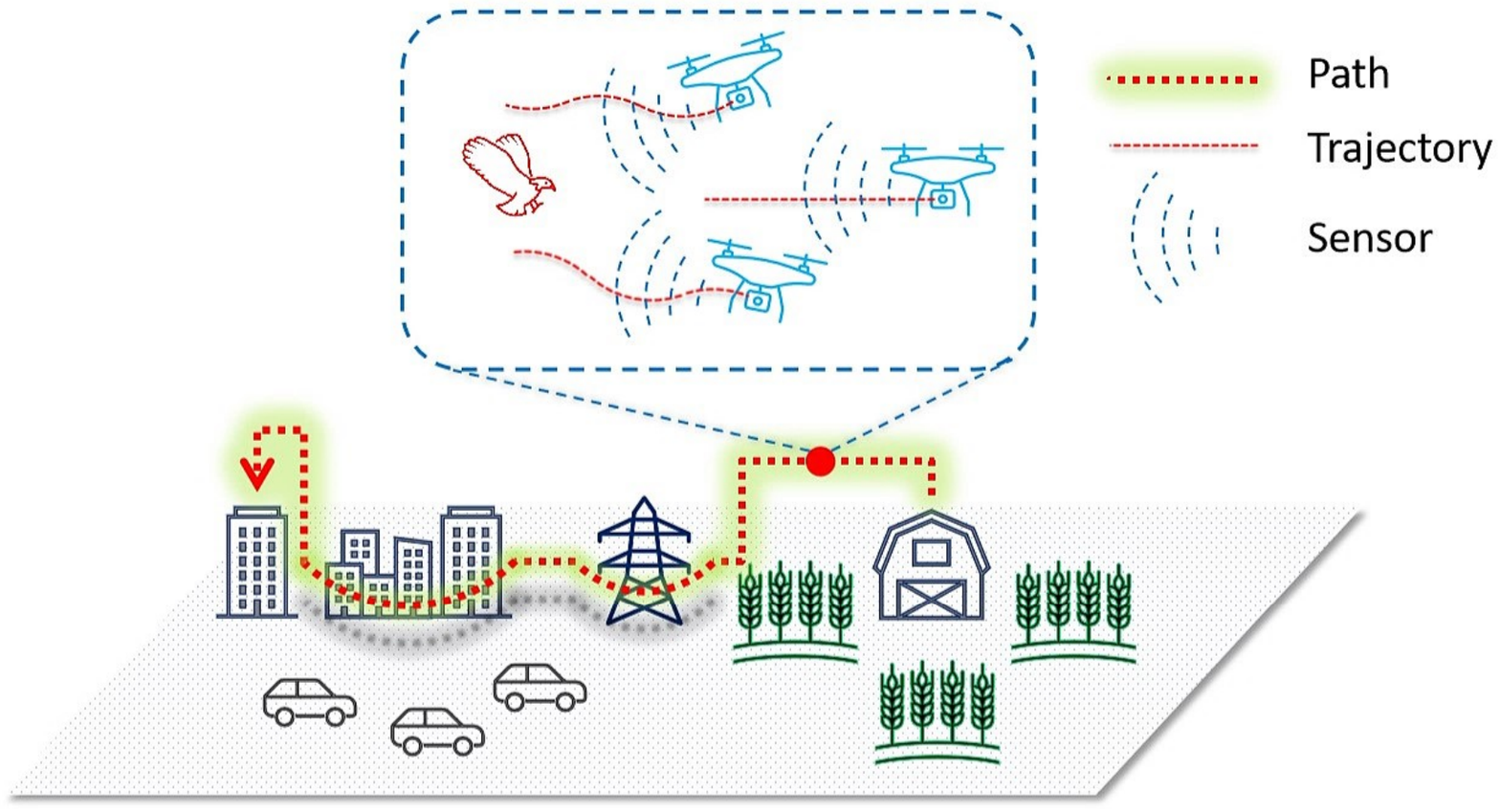} 
	\caption{Scenario of collision avoidance for UAV swarms. }
	\label{fig:scenario}
\end{figure}

We first define a UAV's path and trajectory. A path is a sequence of waypoints that defines the overall global direction of a UAV's flight. On the contrary, a trajectory is a section of the path that provides local velocity and gesture information for the UAV to stay on the path. A typical scenario of collision avoidance for small-sale UAV swarms is illustrated in Fig. \ref{fig:scenario}, in which three UAVs follow pre-planned static paths defined by sequences of waypoints. During the UAVs' flight, when potential collisions are detected, the UAVs take necessary actions to avoid the collisions. Once the collisions are avoided, the UAVs fly toward the next waypoint on their path to resume their course. These upcoming waypoints, which the UAVs aim to reach after avoiding collisions, are referred to as the target positions of the collision avoidance process. 
In the paper we assume that static obstacles such as buildings and towers are avoided in path planning. Our main focus is avoiding collisions with dynamic obstacles such as birds or adversarial UAVs. For energy considerations, we confine the UAVs' movement in two-dimensional space as altitude changes consume much more energy than level flight \cite{huang2021}. 

During collision avoidance, UAVs can plan trajectories to ensure a safe distance from obstacles and between each other, but must return to their original pre-planned paths after the avoidance to resume the mission. We assume the speed of each UAV remains constant during collision avoidance, and each UAV adjusts its direction of movement to void collisions for energy efficiency. This assumption stems from the recognition that consumer- and industry-level UAVs, such as DJI Matrice and Mavic, typically fly at a speed around 10 $m/s$, and therefore dynamic obstacle avoidance usually occurs within a short timeframe. Such a short timeframe makes sudden speed changes impractical due to the large accelerations involved, resulting in a substantial increase in energy consumption. 

It is also assumed that each UAV knows its position, velocity, and that of other swarm members and obstacles within its sensing range, which can be achieved using GPS and LiDAR sensors. 
Typical LiDAR sensors such as SICK MRS1000 are capable to detect obstacles in three-dimensional space with a horizontal aperture angle of 270$^{\circ}$ and a vertical aperture angle of 7.5$^{\circ}$. 
Moreover, onboard microcomputer such as Raspberry Pi is used for decision-making. 

\section{System Model} \label{System}

The problem of multi-UAV collision avoidance is modeled as a Dec-POMDP. This modeling strategy arises from the constraints imposed by the limited aperture angle and sensing range of LiDAR sensors, resulting in a partially observable environment from the UAVs' perspective. The key components of the Dec-POMDP are defined as follows. 

\subsection{Observation} 

The observation of a UAV is defined as a $V\times \textit{4}\times\textit{2}$ array. $V$: $V$ is the maximum number of objects detected, including the ego-UAV. The ego-UAV is always listed in the first row of the observation array. \textit{4}: Each UAV observes four kinetics features, including current position, current velocity, target position, and original velocity. The original velocity of a UAV is its instant velocity before avoiding collisions, which it tries to regain after avoidance maneuvers. Target position and original velocity are only valid for the ego-UAV and are all zeros for other objects, as a UAV only knows the target position and original velocity of its own. \textit{2}: The UAVs operate in a two-dimensional $(x-, y-)$ space. The $x-$ and $y-$ components of the kinetics features are described in two layers of the array. 

\subsection{Action}
\begin{figure}[t] 
	\centering 
	\includegraphics[width=.5\linewidth]{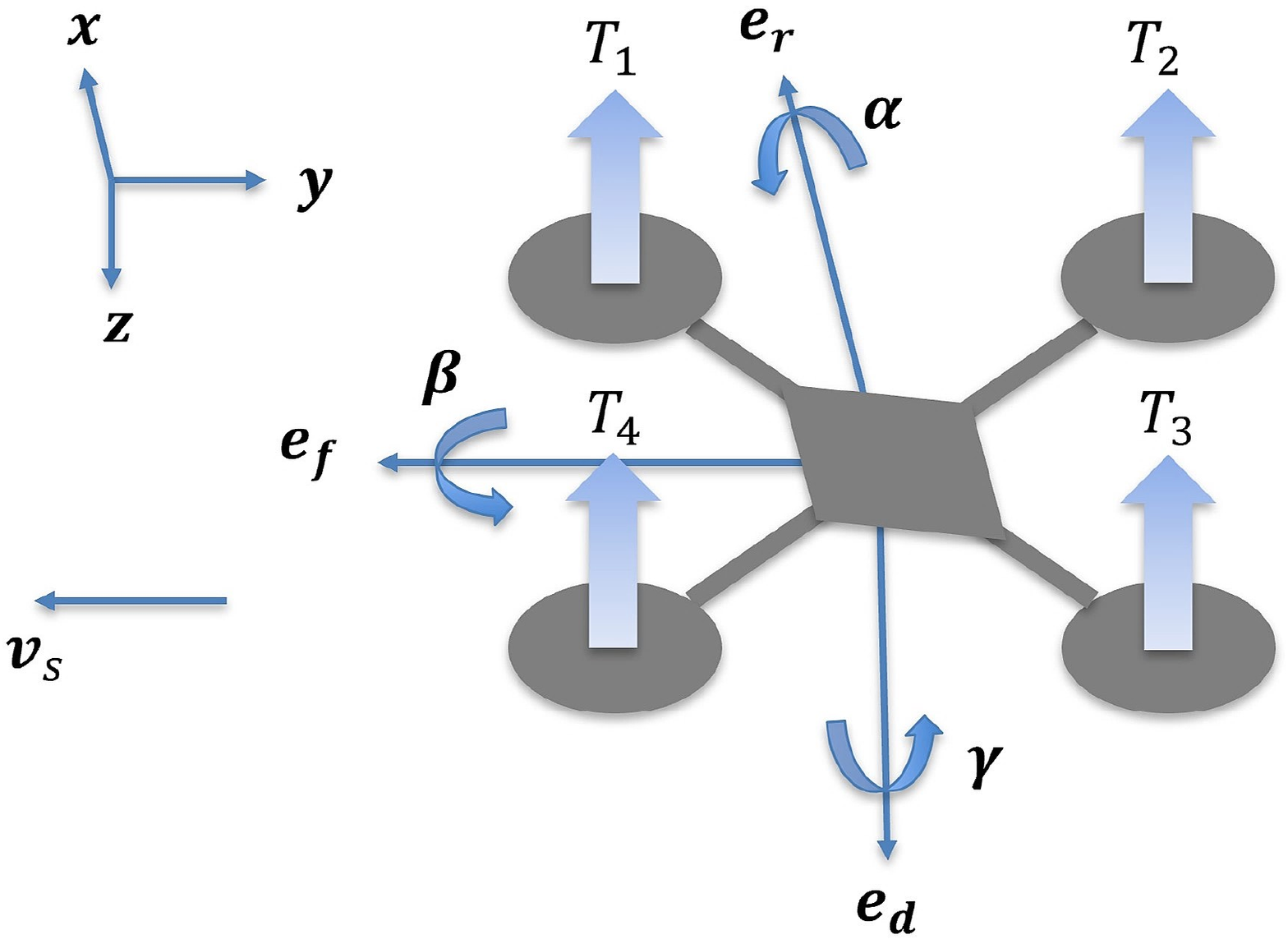} 
	\caption{Motion angles of quad-rotor aircraft. } 
	\label{frame} 
\end{figure}

For simplicity, the action of a UAV is defined by the change in velocity direction as the collision avoidance is performed in two-dimensional space. Nevertheless, it is easy to extend the collision avoidance to three-dimensional space by defining the action of a UAV as its motion angles {Roll} $\alpha$, {Pitch} $\beta$, {Yaw} $\gamma$, and lifting forces $T$ as illustrated in Fig. \ref{frame}, where $\{\boldsymbol{e}_r, \boldsymbol{e}_f, \boldsymbol{e}_d\}$ and $\{\boldsymbol{x}, \boldsymbol{y}, \boldsymbol{z}\}$ are its self and world coordinates, respectively. The UAV's velocity is denoted as $\boldsymbol{v}_s$. To avoid radical velocity change, the action of a UAV is limited in $[-90^{\circ}, 90^{\circ}]$. 

\subsection{Reward} 
The reward signal is designed to encourage the UAVs to successfully execute their mission, avoid collisions, and operate efficiently to save energy. As analyzed in work \cite{huang2021}, saving energy for UAVs can be achieved by ensuring the smoothness of their trajectories. In this paper, smoothness of trajectories is ensured by minimizing velocity changes in the reward signal. 

With the analysis above, two rewards $r_{ave}$ and $r_{min}$ are defined as follows.  
\begin{equation} \label{r_vel}
	\begin{gathered} 
		r_{ave} = \dfrac{1}{n} \sum_{i=1}^{n}r_i + r_{collide} + r_{target}, \\ 
		r_{min} = \min_{i=1, \cdots, n}r_i + r_{collide} + r_{target}, \\ 
	\end{gathered} 
\end{equation} 
where $r_i$ is the reward for UAV $i$ in the swarm, $r_{collide}$ is the penalty for collision and $r_{target}$ is the reward for reaching the target position. These terms are defined as follows.  
\begin{equation*} 
	\begin{aligned} 
		&r_i = \langle|\boldsymbol{v}_s^i|, |\boldsymbol{v}_{tar}^i|\rangle, \\ 
		&r_{collide} = \left\{
		\begin{array}{cl}
			-10 & \text{\textit{if collision}} \\ 
			0 & \text{\textit{otherwise} } \\  
		\end{array}
		\right., \\ 
		&r_{target} = \left\{
		\begin{array}{cl}
			10 & \text{\textit{if mission accomplished}} \\ 
			0 & \text{\textit{otherwise} } \\  
		\end{array}
		\right., 
	\end{aligned} 
\end{equation*} 
where $\langle\cdot, \cdot\rangle$ is inner product, $|\boldsymbol{v}|$ is normalized velocity. $r_i$ measures the deviation of the direction of a UAV's real-time velocity from its original velocity. The value of $r_i$ ranges in $[-1, 1]$, with $r_i=1$ indicating the UAV flies directly along its original velocity, and $r_i=-1$ indicating the UAV flies opposite its original velocity. On the other hand, $r_{collide}$ penalizes actions causing collisions and $r_{target}$ awards actions reaching target position. It just requires $r_{collide}$ and $r_{target}$ to be a very small negative value and a very large positive value compared to $r_i$, respectively. Therefore, empirical values of -10 and 10 are selected. 

As their names suggest, $r_{ave}$ evaluates the swarm by the average performance of all UAVs. While $r_{min}$ evaluates the swarm by the UAV that performs the poorest. The UAV that performs the poorest determines the maximum capacity of the swarm. The rewards $r_{ave}$ and $r_{min}$ can be used in different applications. For example, $r_{ave}$ can be used in applications like communication relay, area coverage, and video streaming, where the average capacity of the swarm, like bandwidth, area size, and data rate, is of more interest. In contrast, $r_{min}$ can be used in applications like unmanned delivery because the successful completion of a mission depends on the poorest performing UAV. 

\section{Algorithm} \label{Algo} 
\subsection{Cooperative Policy Gradient}
\begin{algorithm}
	\caption{Cooperative Policy Gradient}\label{algo1}
	\begin{algorithmic}[1]
		\STATE \textbf{Input}: Initial parameters $\theta$ for actors and $\omega$ for the centralized critic; a temporary replay buffer $R^{tmp}$ and an episodic replay buffer $R^{ep}$. 
		\STATE Initialize $R^{tmp}\leftarrow\emptyset$, $R^{ep}\leftarrow\emptyset$. 
		\FOR{$ep$ in 0,1,2, ...}
		\STATE $t=0$, reset $(s_t, r_t, \boldsymbol{o}_t, \text{done})$. 
		\WHILE{not \text{done}}
		\STATE Sample actions $\{\boldsymbol{a}_t, a^i_t\sim\pi_{\theta_t}(\cdot|o^i_t)\}$. 
		\STATE Get $(s_{t+1}, \boldsymbol{o}_{t+1}, r_t, \text{done})$ from the environment. 
		\STATE $R^{tmp}\leftarrow(s_t, \boldsymbol{o}_t, \boldsymbol{a}_t, r_t, s_{t+1}, \boldsymbol{o}_{t+1}, \text{done})$
		\STATE $t += 1$
		\ENDWHILE
		\STATE $R^{ep}\leftarrow R^{tmp}$, $R^{tmp}\leftarrow\emptyset$. 
		\STATE Sample batches $\mathcal{B}$ from $R^{ep}$. 
		\STATE Get the joint action value $Q_{tot} = f_{\omega}(s_t, \boldsymbol{o}_t, \boldsymbol{a}_t)$. 
		\STATE Get the advantage $A_i=Q_{tot}-b_i$ for each agent based on {Algorithm} \ref{algo2}. 
		\STATE Update actors for each agent by the policy gradient in Eq. \eqref{policygradient}. 
		\STATE Update critic by the temporal loss in Eq. \eqref{criticloss}. 
		\ENDFOR
	\end{algorithmic}
\end{algorithm}

\begin{algorithm}
	\caption{Counterfactual Baseline}\label{algo2}
	\begin{algorithmic}[1]
		\STATE \textbf{Input}: For time step $t$, get centralized critic $f_{\omega}(s_t, \boldsymbol{o}_{t}, \boldsymbol{a}_t)$, joint observation $\boldsymbol{o}_t$ and joint action $\boldsymbol{a}_t$ of all agents, agent ID $\nu\equiv\{0, 1, ..., n-1\}$. 
		\FOR{each agent $i\in\nu$}
		\FOR{$j$ in $\nu\setminus i$} 
		\STATE Get coalition $\boldsymbol{U}^{-j}$ by excluding agent $j$ from $\nu$. 
		\STATE Sample $\tilde{\boldsymbol{a}}$ from $\mathcal{A}$ uniformly. 
		\FOR{each $\tilde{a}_l\in\tilde{\boldsymbol{a}}$}
		\STATE Get $\tilde{Q}_{j,l} = f_{\omega}(s_t, \boldsymbol{o}_t, \boldsymbol{a}^{-j}_t(a_i\leftarrow\tilde{a}_l))$. 
		\ENDFOR
		\ENDFOR
		\STATE Get the baseline $b_i = \mathbb{E}_{j,l}[\tilde{Q}_{j,l}]$. 
		\ENDFOR
	\end{algorithmic}
\end{algorithm}

The cooperative policy gradient in our MARL algorithm can be summarized in {Algorithm} \ref{algo1}. The parameters of actors, centralized critic, and two experience buffers $R^{tmp}$ and $R^{ep}$ are first initialized (Line 1-2). $R^{tmp}$ is for storing state-action pairs within one episode and is reset after each episode. $R^{ep}$ is for storing episodic state-action pairs, which are copied from $R^{tmp}$ each episode (Line 3-11). To update the actor networks, the joint action value is first estimated by the centralized critic (Line 13). 
After that, an advantage function is derived for each agent by subtracting a counterfactual baseline $b_i$ from the joint action value $Q_{tot}$ (Line 14). 
The actors and critic are updated by policy gradient and temporal loss, respectively (Line 15-16). It is noted that policy $\pi_{\theta_i}$ is deterministic for continuous actions. 
More details on deriving the counterfactual baseline are given below. 

\subsection{Counterfactual Baseline} 

The counterfactual baseline is derived following {Algorithm} \ref{algo2}. Agent $i$ is first grouped into $n$ different coalitions $\boldsymbol{U}_{i}\equiv\{... ,\boldsymbol{U}^{-j},...\}$. Each coalition excludes agent $j, (j\in\{0,\cdots,n-1\}, j\neq i)$ by masking out $a_j$ from the joint actions (Line 5). For coalition $\boldsymbol{U}^{-j}$, $k$ actions are sampled from the action space $\mathcal{A}$ uniformly for agent $i$ (Line 6). For each sampled action $\tilde{a}_l$, a counterfactual joint action $\tilde{\boldsymbol{a}}_t$ is derived by replacing $a_i$ with $\tilde{a}_l$ in $\boldsymbol{a}_t$ while keeping other agents' actions unchanged. Then, a counterfactual action value $\tilde{Q}$ for each sample action in each coalition is acquired (Line 8). Eventually, the baseline for agent $i$ is defined by the expectation of all counterfactual action values (Line 11). 

\begin{figure}[t]
	\centering
	\includegraphics[width=0.5\textwidth]{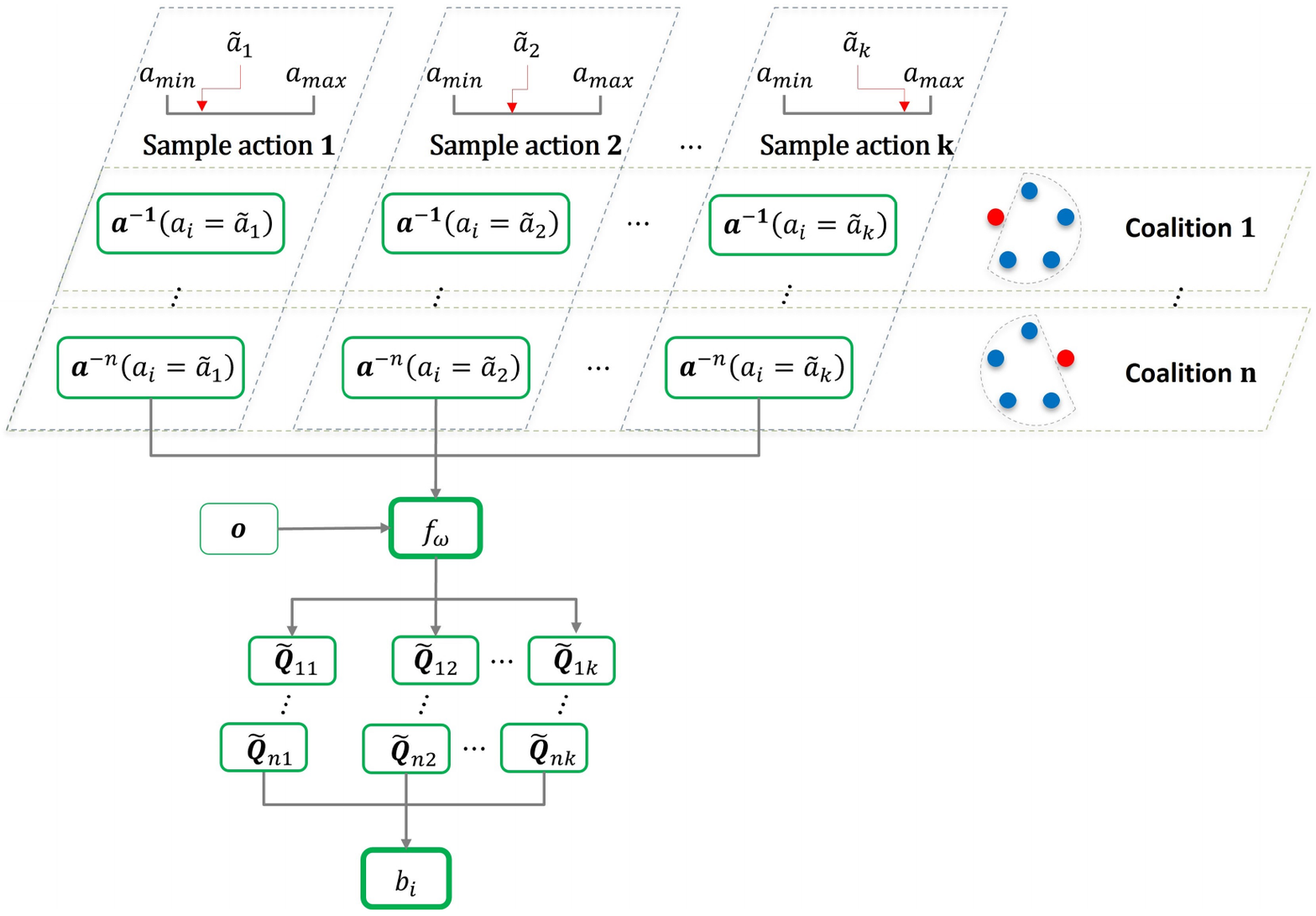} 
	\caption{Illustration of the counterfactual baseline. }
	\label{baseline}
\end{figure}

The counterfactual baseline is further illustrated with Fig.~\ref{baseline}. As depicted, the counterfactual baseline is constructed in two dimensions: action and the coalition of agents. {This two-dimensional approach is expected to enable our method to offer a more precise estimation and allocation of credits to agents when compared to existing MARL algorithms which only consider either the actions of agents like COMA or the interrelation between agents like Shapley Q learning. }
On the other hand, compared to Shapley Q learning whose computational complexity is $N!$, our method only has a complexity of $N\times K$, where $N$ and $K$ represent the number of agents in the swarm and the number of sampled actions in calculating the baseline. Moreover, the derivation of our baseline did not require any assumption on the value functions. Source codes are available at request. 

\subsection{Convergence Proof} 
Here, we prove our baseline does not affect the convergence of deterministic policy gradient. 
Previous work on single agent actor-critic \cite{suttonpg} has shown that a baseline $b$ does not affect the convergence of stochastic policy gradient if $b$ is action-independent. COMA further proves that in multi-agent scenarios, a stochastic policy gradient is not affected if baseline $b_i$ is $a_i$ independent. Can this conclusion be generalized to a deterministic policy gradient? The authors in \cite{dpg} prove that a deterministic policy $\pi_{\theta}$ is a special case of stochastic policy by writing a stochastic policy as $\mu_{\pi_{\theta}, \sigma}$, where $\sigma$ is a variance parameter, such that $\mu_{\pi_{\theta}, 0}\equiv\pi_{\theta}$. Using this expression and following similar steps as in COMA, {Lemma} \ref{lemma1} is given as follows. 
Detailed derivations are available at request. 
\begin{lemma} \label{lemma1}
	For a policy gradient at step $t$: 
	\begin{equation} \label{policygradient1} 
		\begin{gathered} 
			g_t = \mathbb{E}_{\mu} \left[\sum_{i}\nabla_{\theta}\log\mu_{\theta, \sigma}(a_i|\tau_i)\cdot A_i
			\right], \\ 
			A_i = Q - b_i, 
		\end{gathered} 
	\end{equation}
	where $b_i$ is an agent-specific baseline defined in {Algorithm} \ref{algo2}, $b_i$ does not introduce any bias to the action value, and hence does not affect the convergence of the policy gradient. 
\end{lemma}

\begin{proof} 
	We need to only look at the expectation of the contribution of the baseline $b_i$ with regard to the current policy in Eq. (\ref{policygradient1}). If the expectation of $b_i$ is 0, Eq. (\ref{policygradient1}) has no bias in convergence. 
	Let the joint policy be: $\boldsymbol{\mu} = \prod_{i}\mu_i(a_i|\tau_i)$. Then the proof for this lemma can follow the same logic as for the single agent actor-critic algorithm. Let $d^{\boldsymbol{\mu}}(s)$ be the discounted state distribution defined as follows: $d^{\boldsymbol{\mu}}(s)= \sum_{t=0}^{\infty}\gamma^tPr(s_t=s|s_0, \boldsymbol{\mu})$, the second part of Eq. (\ref{policygradient1}) is: 
	\begin{equation} \label{gb} \nonumber
		\begin{aligned} 
			&g_b = -\mathbb{E}_{\mu} \left[\sum_{i}\nabla_{\theta}\log\mu_i(a_i|\tau_i)\cdot b_i\right] \\ 
			&= -\sum_{s}d^{\boldsymbol{\mu}}(s) \sum_{i} \sum_{\boldsymbol{a}^{-i}}\boldsymbol{\mu}^{-i}\cdot  
			\sum_{{a}^{i}} \mu_i(a_i|\tau_i) \nabla_{\theta}\log\mu_i(a_i|\tau_i) b_i \\ 
			&= -\sum_{s}d^{\boldsymbol{\mu}}(s) \sum_{i} \sum_{\boldsymbol{a}^{-i}}\boldsymbol{\mu}^{-i}\cdot
			b_i \sum_{{a}^{i}} \mu_i(a_i|\tau_i) \nabla_{\theta}\log\mu_i(a_i|\tau_i) \\ 
			&(b_i \text{ is not a function of }a_i) \\
			&= -\sum_{s}d^{\boldsymbol{\mu}}(s) \sum_{i} \sum_{\boldsymbol{a}^{-i}}\boldsymbol{\mu}^{-i}\cdot 
			b_i \nabla_{\theta}\sum_{{a}^{i}}\mu_i(a_i|\tau_i)\\ 
			&= -\sum_{s}d^{\boldsymbol{\mu}}(s) \sum_{i} \sum_{\boldsymbol{a}^{-i}}\boldsymbol{\mu}^{-i}\cdot 
			b_i \nabla_{\theta}1\\
			&= 0. 
		\end{aligned} 
	\end{equation}
	As shown above, the baseline doesn't introduce any bias to the action value. Hence, the baseline $b_i$ does not affect the convergence of policy $\mu$ and $\mu\equiv\pi$ when $\sigma=0$. 
\end{proof}

\section{Experiments} \label{Exper}
\subsection{Environment Design} 
\begin{figure}[t] 
	\centering 
	\includegraphics[width=.6\linewidth]{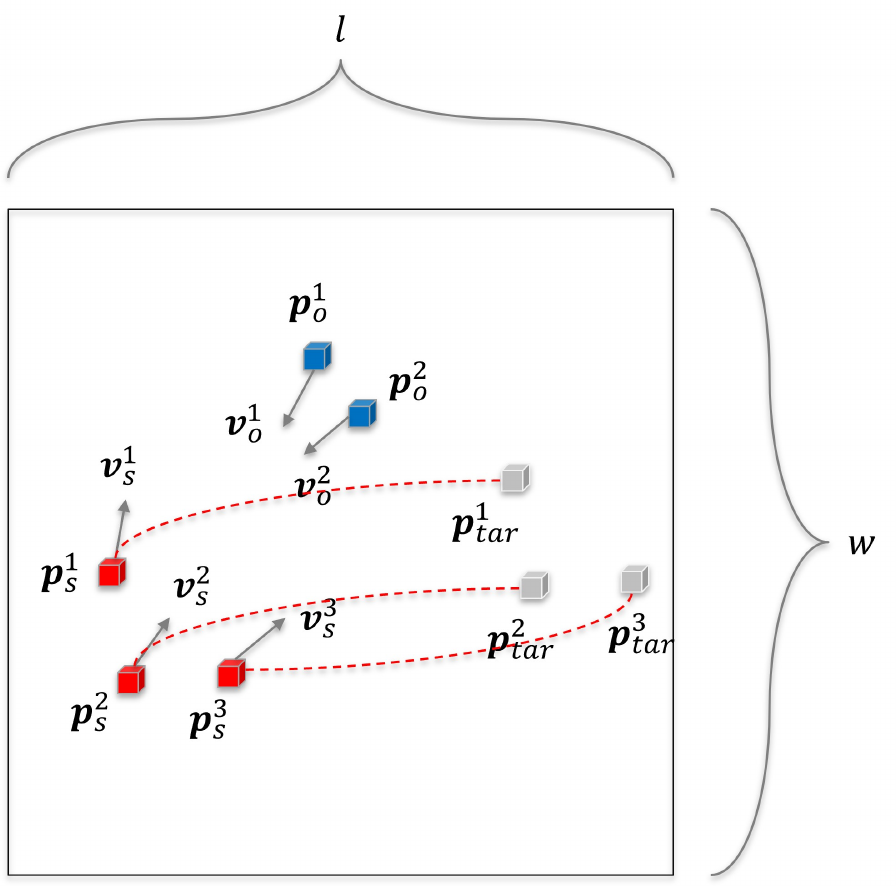} 
	\caption{Illustration of the environment. } 
	\label{env} 
\end{figure} 

We design our own \textit{Gym}-like environment to best simulate the cooperative collision avoidance of UAVs as existing MARL environments cannot closely simulate multi-UAV applications. The environment is developed based on the application scenario described in Section \ref{Appli} and is illustrated in Fig.~\ref{env}. Both the UAVs and obstacles are dynamic and moving in a square with width $w$ and length $l$. In the illustration, the UAVs and obstacles are marked with red and blue cubes, respectively. Velocities of the UAVs and obstacles are depicted with grey arrows. The target positions of UAVs are depicted as gray cubes. The pre-planned trajectories of UAVs to their respective target positions are shown in dashed curves. 

The environment operates in an episodic manner. In each episode, the mission starts when the UAVs and obstacles are spawned on the left and right side of the square, respectively. The UAVs fly along their pre-planned trajectories and change the directions of their velocities to avoid collisions. Collisions occur when the distance between any UAV and obstacle is smaller than a threshold distance $d_{obs}$, or the distance between any two UAVs operating in the swarm is smaller than $d_{v2v}$. 
The episode terminates when collisions occur, or all UAVs reach their respective target positions. The observation, action and reward at each step are described in Section \ref{System}. 
Source codes are available at request. 

\subsection{Network Design} 
The actor network consists of a Gated Recurrent Unit (GRU) layer with 32 units, preceded and followed by two fully connected layers with 32 neurons and a ReLU activation function, respectively. 
Finally, the output layer gives an action with a $\tanh$ activation function. 
On the other hand, the centralized critic takes three inputs: the joint observation, the joint action, and the state of the environment, each using a fully connected layer with 32 neurons. The inputs are concatenated and passed to a dense layer with 256 neurons, and a ReLU activation function. Finally, an output layer produces a joint action value. The target critic network has the same structure with the centralized critic. 

\subsection{Experiment Setup} 
Actions of UAVs are selected by $\epsilon$-greedy rules: $a = \pi_{\theta}(\cdot|o) \cdot (1-\epsilon) + \mathcal{N} \cdot \epsilon$, where $\mathcal{N}$ is a Gaussian noise.  $\epsilon$ is annealed exponentially from 1.0 to 0.1 across 12$k$ episodes. 
The target critic networks are updated each episode by $\bar{\omega} = \tau\cdot\omega + (1-\tau)\cdot\bar{\omega}$, where $\bar{\omega}$ are parameters of target critic network, $\omega$ are parameters of critic network and $\tau=0.01$. Future rewards are discounted by $\gamma=0.99$. Actors share parameters to boost training. 
On the other hand, the velocities of UAVs and obstacles are set to 10 $m/s$ and 5 $m/s$, respectively, to reflect the typical velocity range of off-the-shelf UAVs, such as DJI Matrice and Mavic. The sensing distance of UAVs is set to 100 $m$. The minimum safety distance to avoid collisions is set to 10 $m$. 

\subsubsection{Compare with MARL Algorithms}
We first compare $CoDe$ with other advanced MARL algorithms such as COMA, Shapley Q Learning, VDN, IQL, and Naive Default to demonstrate the advantage of $CoDe$ in learning cooperative policies.  Naive Default adopts the same framework as $CoDe$, only that the counterfactual baseline is calculated by replacing the agent's action with default action $a=0$, while keeping other agents unchanged. In addition, it is worth noting that COMA is designed for discrete action space. To apply COMA in continuous action space, ten actions are sampled from $[-90^{\circ}, 90^{\circ}]$ uniformly for each agent. Different from all other value-based methods, COMA is a policy-based method and adopts on-policy training. 



In alignment with our predefined assumptions of small-sale UAV swarms, we evaluate the performance of our algorithm in two distinct scenarios: two-UAV-one-obstacle (2U1O) and three-UAV-two-obstacle (3U2O) using the two rewards $r_{ave}$ and $r_{min}$. 
Scenarios with more UAVs will be included in future work as it takes very long for the critic network to converge given the high dimension of the joint observation. 

\subsubsection{Compare with Conventional Algorithms}
Furthermore, we also compared $CoDe$ with $E^2Coop$ \cite{huang2021} to demonstrate the advantage of $CoDe$ on energy efficiency and reaction time over conventional algorithms. $E^2Coop$ utilizes meta-heuristics to plan collision avoidance strategies for UAVs, and demonstrates superior performances in safety and energy efficiency to other conventional algorithms. 

For simplicity, we only test in scenario 3U2O using $r_{ave}$. We acquire the numerical results on safety and energy efficiency using 100 trajectories generated by pre-trained $CoDe$ models and $E^2Coop$, respectively. Unlike $CoDe$, $E^2Coop$ minimizes a cost function to avoid collisions at each step, rather than maximizing reward signals. Specifically, the cost function converts the energy consumption of a UAV to the average curvature of its trajectory. Hence, we employ the average curvature as a metric to assess energy efficiency for $CoDe$ and $E^2Coop$. The average curvature of a trajectory is defined as follow.  
\begin{equation} \label{curvature} 
	\begin{aligned} 
		E_n = e_v \int \mid S''(p) \mid dp, 
	\end{aligned} 
\end{equation} 
where $p$ is arc length parameter, $S(p)=[x(p), y(p)]$ denotes a waypoint on the trajectory, $S''(p)$ denotes the second order derivative of $S(p)$, $e_v$ is a velocity dependent coefficient. 

\begin{figure}[htbp] 
	\centering
	\begin{subfigure}[b]{0.49\linewidth}
		\centering
		\includegraphics[width=1.0\linewidth]{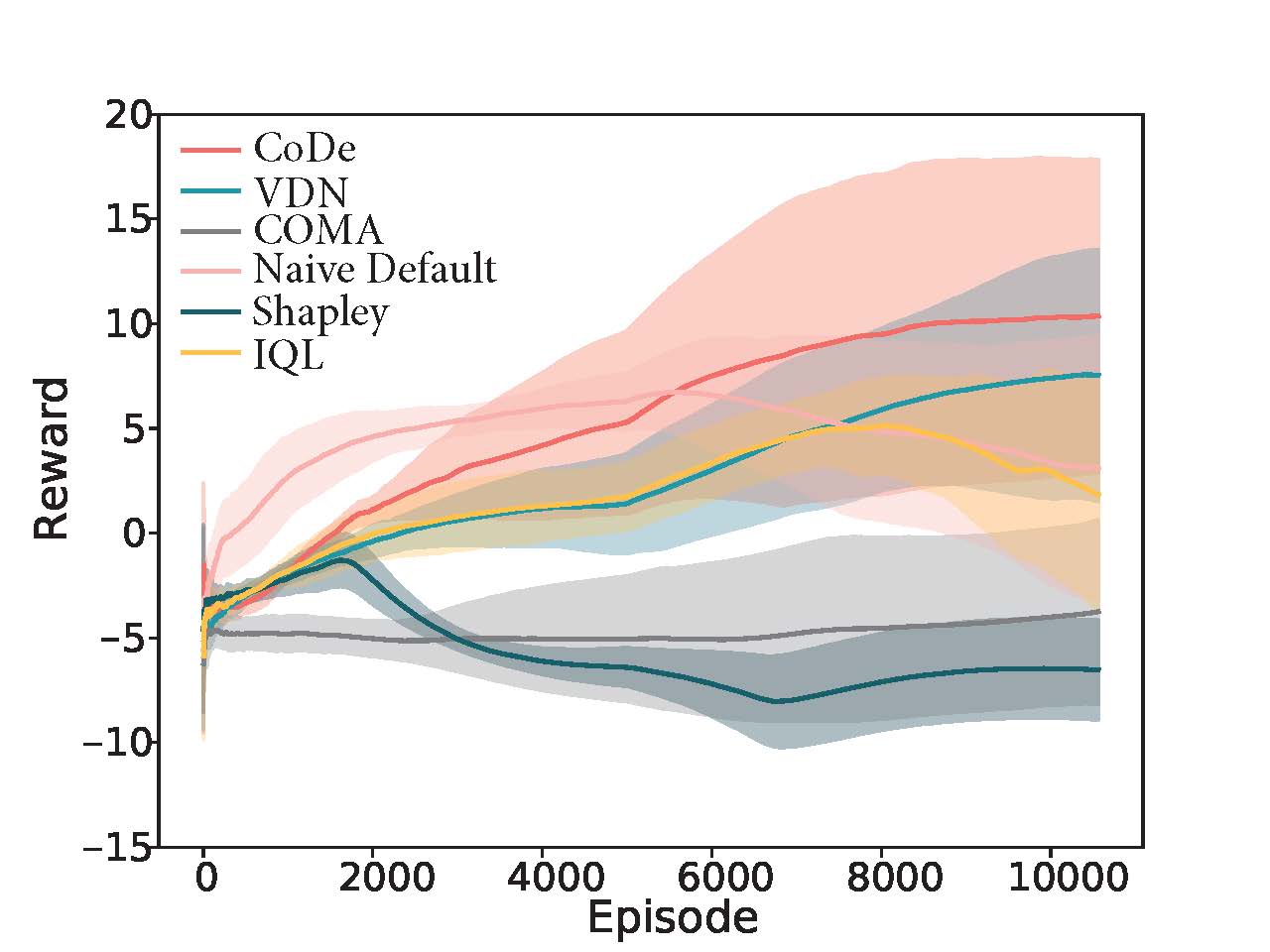} 
		\caption{\textit{2U1O} with reward $r_{ave}$. } 
		\label{5a} 
	\end{subfigure} 
	\begin{subfigure}[b]{0.49\linewidth}
		\centering
		\includegraphics[width=1.0\linewidth]{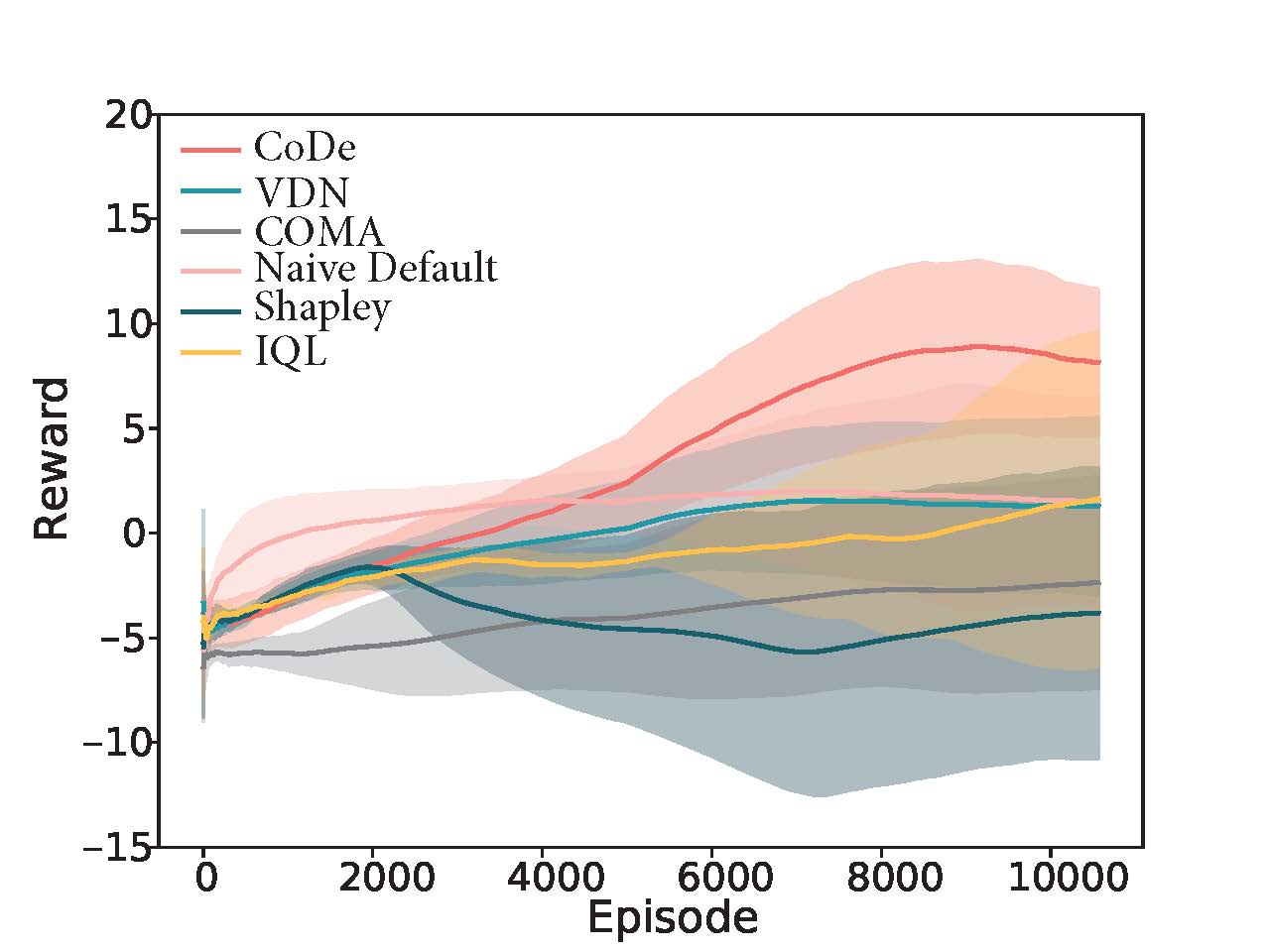} 
		\caption{\textit{3U2O} with reward $r_{ave}$. } 
		\label{5b} 
	\end{subfigure} 
	\begin{subfigure}[b]{0.49\linewidth}
		\centering
		\includegraphics[width=1.0\linewidth]{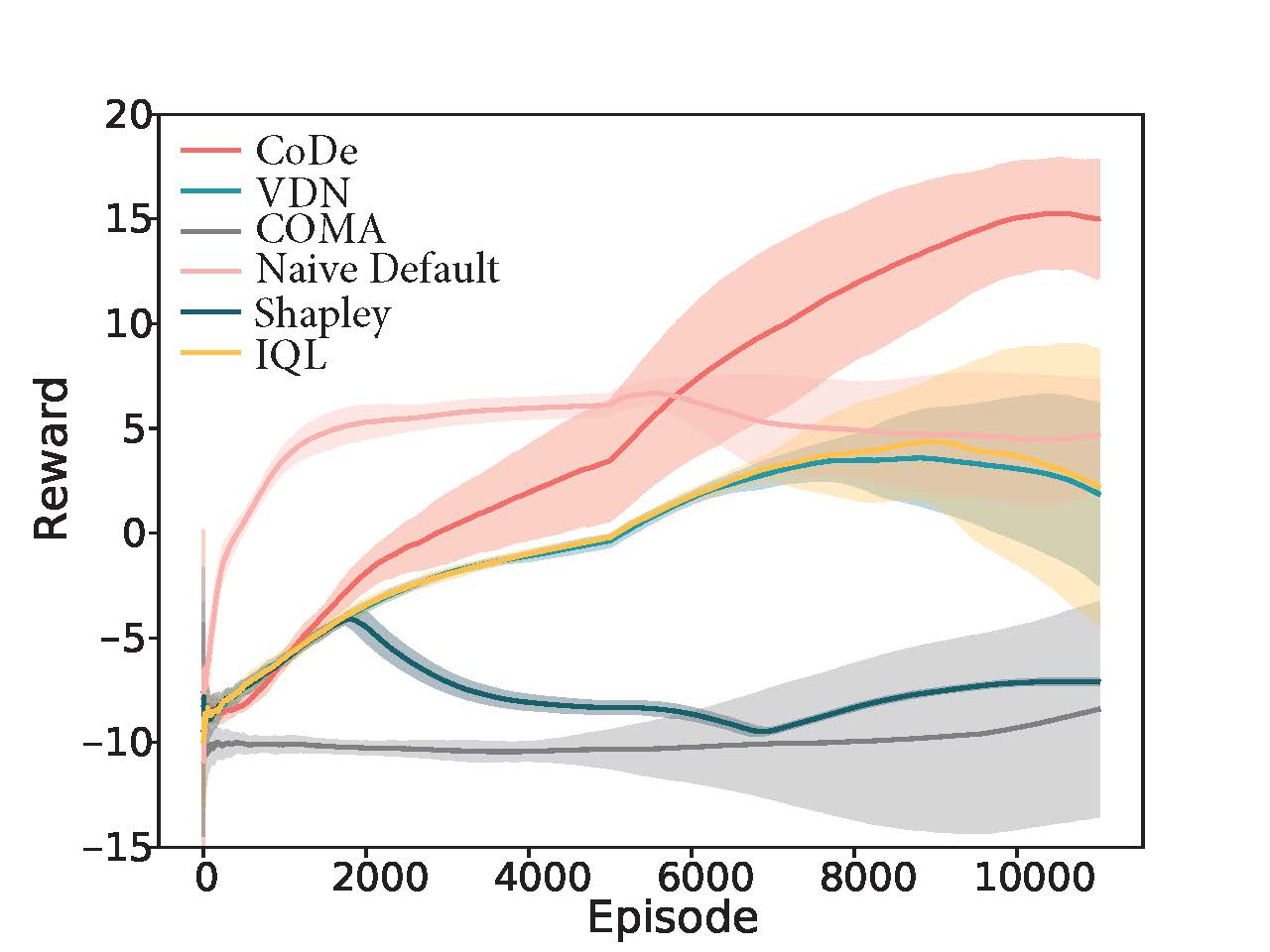} 
		\caption{\textit{2U1O} with reward $r_{min}$. } 
		\label{5d} 
	\end{subfigure} 
	\begin{subfigure}[b]{0.49\linewidth}
		\centering
		\includegraphics[width=1.0\linewidth]{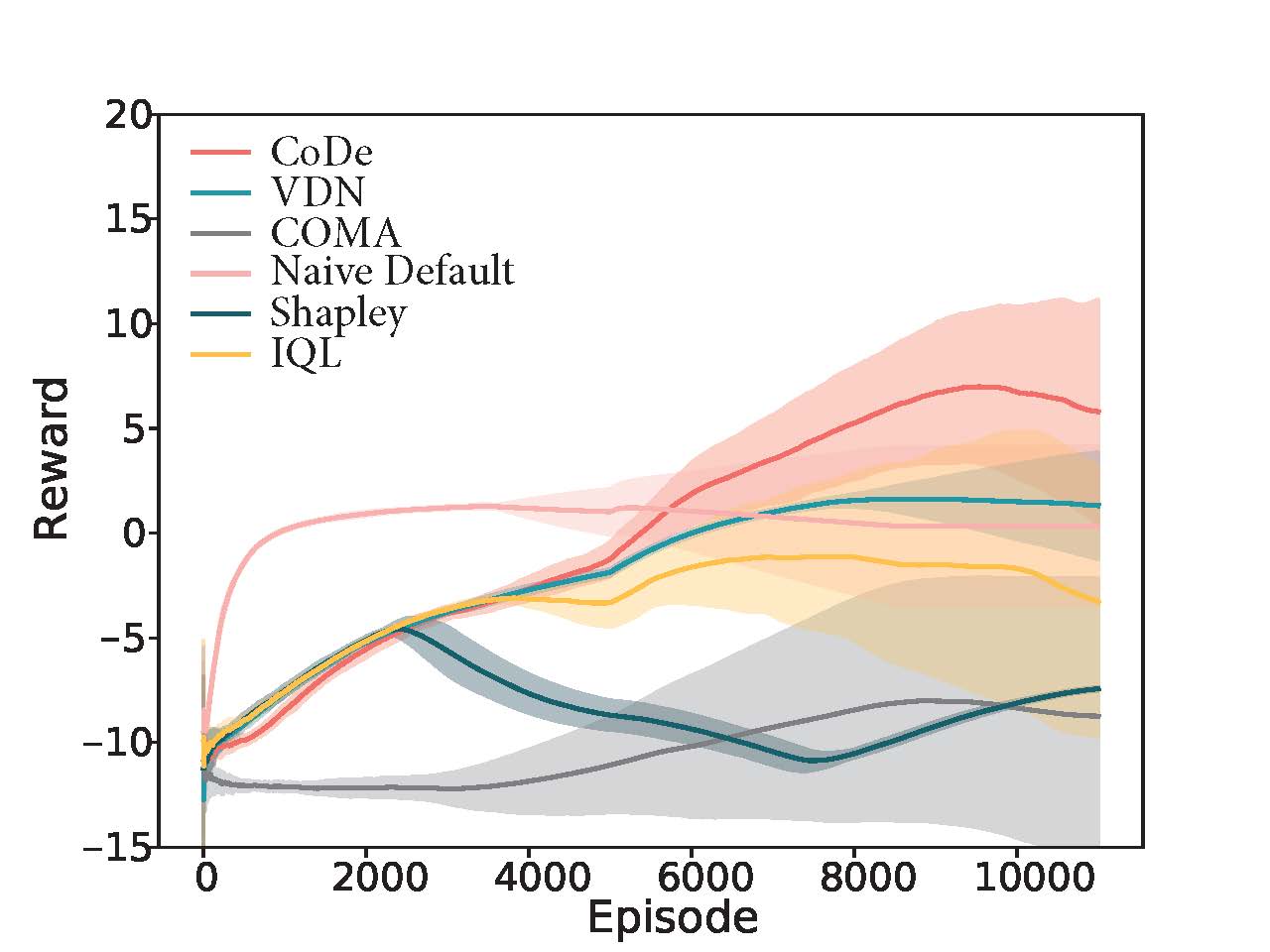} 
		\caption{\textit{3U2O} with reward $r_{min}$. } 
		\label{5e} 
	\end{subfigure} 
	\caption{Learning curves of $CoDe$ and other algorithms. } 
	\label{learningcurves} 
\end{figure} 

\begin{table*}[h]
	\centering
	\begin{tabular}{||c |c c c c c c||} 
		\specialrule{.1em}{.1em}{.1em}
		\multicolumn{7}{|c|}{Average Rewards} \\
		\specialrule{.1em}{.1em}{.1em}
		&  CoDe  &  Naive Default  &  VDN  &  COMA  &  Shapley  &  IQL  \\ 
		\specialrule{.1em}{.1em}{.1em} 
		\textit{Scenario 2U1O with }$r_{ave}$	  & \textbf{10.3548}  & {7.5414}   & -3.7882  & 3.1346   & -6.5211 & 1.9271 
		\\ [0.5ex] 
		& \textbf{$\pm$ 7.4983 } & {$\pm$ 5.9884 } & $\pm$ 4.3642  & $\pm$ 6.2866  & $\pm$ 2.3877  & $\pm$ 5.4299  \\ [0.5ex] 
		\specialrule{.1em}{.1em}{.1em} 
		\textit{Scenario 3U2O with }$r_{ave}$ & \textbf{8.1630}   & {1.5522}   & 1.2905 & -2.3754    & 1.6017   & -3.8316 \\ [0.5ex] 
		& \textbf{$\pm$ 3.5079 } & {$\pm$ 4.9130 } & $\pm$ 4.2112  & $\pm$ 5.0504  & $\pm$ 6.9373  & $\pm$ 8.0040  \\ [0.5ex] 
		\specialrule{.1em}{.1em}{.1em}
		\textit{Scenario 2U1O with }$r_{min}$	  & \textbf{15.0234}  & {4.6630}   & 1.9819 & -8.4646 & -7.0930 & 2.3201 
		\\ [0.5ex] 
		& \textbf{$\pm$ 2.7736 } & {$\pm$ 2.6657 } & $\pm$ 4.2387  & $\pm$ 5.0645  & $\pm$ 0.1360  & $\pm$ 6.4832  \\ [0.5ex] 
		\specialrule{.1em}{.1em}{.1em}
		\textit{Scenario 3U2O with }$r_{min}$ & \textbf{5.8231}   & {0.3393}   & 1.3205 & -8.7307    & -7.4533 & -3.2055  \\ [0.5ex] 
		& \textbf{$\pm$ 5.2555 } & {$\pm$ 3.7980 } & $\pm$ 2.5263  & $\pm$ 6.5867  & $\pm$ 0.0911  & $\pm$ 6.4764  \\ [0.5ex] 
		\specialrule{.1em}{.1em}{.1em}
	\end{tabular}
	\caption{Average rewards and standard deviations over the last 100 training episodes. }
	\label{table1}
\end{table*}

\begin{table*}[h]
	\centering
	\begin{tabular}{||c |c c c c c||} 
		\specialrule{.1em}{.1em}{.1em}
		\multicolumn{6}{|c|}{\textit{Scenario 3U2O with }$r_{ave}$} \\
		\specialrule{.1em}{.1em}{.1em}
		& Reaction & Energy & $d_{U2O}$ & $d_{U2U}$ & Collision \\ 
		& Time (Sec) & Cost &  &  & Rate \\ [0.5ex] 
		\specialrule{.1em}{.1em}{.1em}
		$E^2Coop$ 		   & 0.48 $\pm$ 0.05  & 134.88 $\pm$ 8.3  & 38.70 & 40.31 & 0 \\ [0.5ex]  
		\textit{CoDe}  & \textbf{0.007 $\pm$ 0.01 } & \textbf{75.93 $\pm$ 31.8 } & \textbf{19.83} & \textbf{28.26} & \textbf{3\%} \\ [0.5ex] 
		\specialrule{.1em}{.1em}{.1em}
	\end{tabular}
	\caption{Numerical results on safety and energy efficiency of $CoDe$ and $E^2Coop$. }
	\label{table2}
\end{table*}

\subsection{Experiment Results} 

\subsubsection{Compare with MARL Algorithms}
The safety and energy efficiency of MARL algorithms are reflected in the achieved average rewards, as our reward function includes both the velocity change of UAVs and a collision penalty. By maximizing the average reward, the UAV minimizes its energy consumption on top of ensuring safety. Therefore, we believe it is sufficient to focus on learning performances, such as convergence speed, average reward, and standard deviation when comparing between MARL algorithms. These performances directly reflect the quality of credit assignment schemes, in terms of how much information the agents can learn from the joint action value, and how useful this information is to individual agent. 

{Fig. \ref{learningcurves} gives the average learning curves and variations of all the algorithms. The results show that $CoDe$ achieves the highest average score in all the scenarios. Especially, when the scenario gets complex with three UAVs and two obstacles (\textit{3U2O}), only $CoDe$ successfully finds cooperative policies, and all other algorithms fail to converge. This is because of the advantage brought by the novel credit assignment scheme of $CoDe$ in training cooperative policies. On the other hand, VDN achieves the second best performance in scenario \textit{2U1O} using $r_{ave}$, and fails to converge in scenario \textit{2U1O} using $r_{min}$. This is because the reward $r_{ave}$ matches the additive assumption on value functions in VDN, and $r_{min}$ does not. Specifically, Shapley Q Learning and COMA fail to find cooperative policies in any scenario, simply due to the high computational complexity in rendering Shapley values and the inefficiency of COMA's on-policy training. In addition to the learning curves, the average rewards and standard deviations of the last 100 training episodes are provided in Table \ref{table1}. The numerical results show that $CoDe$ achieves higher average scores than existing MARL algorithms. } 

\subsubsection{Compare with Conventional Algorithms} 

The results on safety and energy efficiency of $CoDe$ and $E^2Coop$ are shown in Table \ref{table2}. 
The distances presented in Table \ref{table2} are the minimum distances between UAVs and obstacles ($d_{U2O}$) and among UAVs ($d_{U2U}$) before collisions occur. This is because an episode terminates when collisions occur in our simulation environment, after which a new episode is initiated.  
The results show that the reaction time of $CoDe$ improves over 90\% compared to $E^2Coop$. This is because decisions in $CoDe$ are made by a simple forward pass of neural networks, rather than by environment modeling and heuristic search as in $E^2Coop$. 
On the other hand, the energy consumption is reduced by 43.71\% using $CoDe$, compared to $E^2Coop$. This is because $CoDe$ is able to find shorter and smoother trajectories than $E^2Coop$. The minimum $d_{U2O}$ and $d_{U2U}$ are larger than the safety distance for all three algorithms in both scenarios. However, like all other machine learning-based methods, $CoDe$ has a failure rate when deployed online which causes collision rates of UAVs. The results show that $CoDe$ has a collision rate of 3\% compared to zero of $E^2Coop$. 

\subsection{Trajectory Demonstration}
\begin{figure}[htbp] 
	\centering
	\begin{subfigure}[b]{0.46\linewidth}
		\centering
		\includegraphics[width=1.0\linewidth]{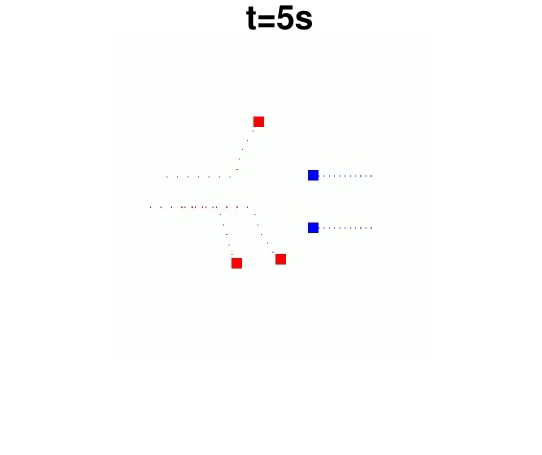} 
		\caption{$E^2Coop$ (t=5$s$). } 
		\label{demo-a} 
	\end{subfigure} 
	\begin{subfigure}[b]{0.46\linewidth}
		\centering
		\includegraphics[width=1.0\linewidth]{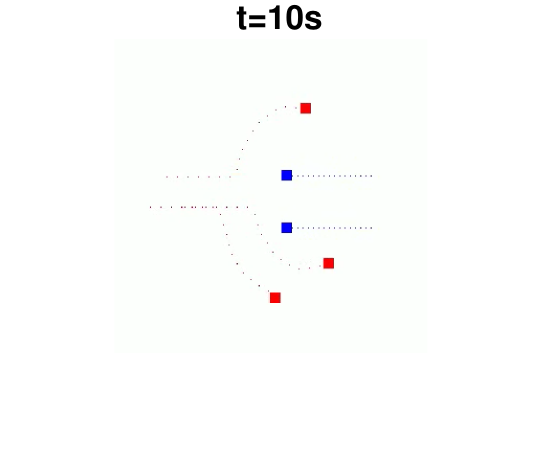} 
		\caption{$E^2Coop$ (t=10$s$). } 
		\label{demo-b} 
	\end{subfigure} 
	\begin{subfigure}[b]{0.46\linewidth}
		\centering
		\includegraphics[width=1.0\linewidth]{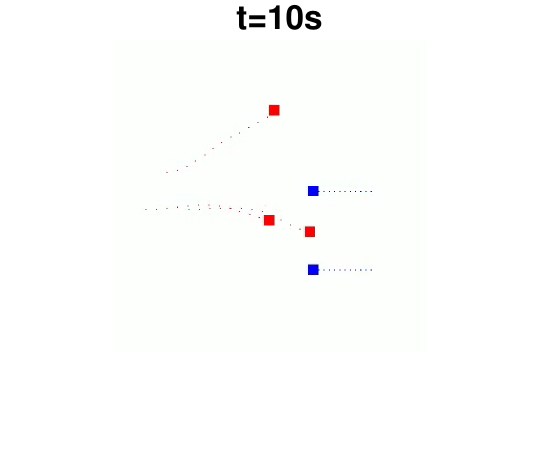} 
		\caption{$CoDe$ (t=10$s$). } 
		\label{demo-c} 
	\end{subfigure} 
	\begin{subfigure}[b]{0.46\linewidth}
		\centering
		\includegraphics[width=1.0\linewidth]{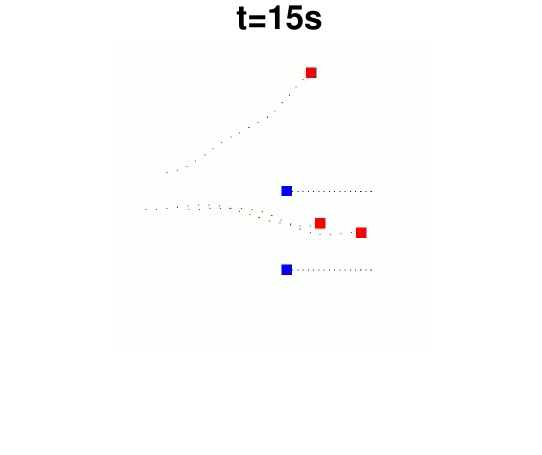} 
		\caption{$CoDe$ (t=15$s$). } 
		\label{demo-d} 
	\end{subfigure} 
	\caption{Trajectories generated by $E^2Coop$ and $CoDe$. }
	\label{demo} 
\end{figure} 
The trajectories of UAVs using $E^2Coop$ and $CoDe$ are demonstrated with dashed line in Figure \ref{demo}. 
It is observed that $CoDe$ generates smoother trajectories for UAVs to navigate around obstacles, in contrast to those generated by $E^2Coop$ that tend to circumvent obstacles. Although UAVs are closer to each other using $CoDe$, they are considered safe and efficient from a long-term perspective, as it's easier for them to get back to their original formation.



\section{Conclusion} \label{Concl}
In this paper, we propose an innovative collision avoidance algorithm for small-sale UAV swarms using MARL.
Our algorithm addresses the challenges of effective cooperation and optimizing energy efficiency of UAVs in collision avoidance through an innovative credit assignment scheme. 
Extensive experiments are conducted. 

\addtolength{\textheight}{-12cm}   



%
%



\bibliographystyle{IEEEtran}
\bibliography{IEEEfull}

\begin{thebibliography}{10}
\providecommand{\url}[1]{#1}
\csname url@rmstyle\endcsname
\providecommand{\newblock}{\relax}
\providecommand{\bibinfo}[2]{#2}
\providecommand\BIBentrySTDinterwordspacing{\spaceskip=0pt\relax}
\providecommand\BIBentryALTinterwordstretchfactor{4}
\providecommand\BIBentryALTinterwordspacing{\spaceskip=\fontdimen2\font plus
\BIBentryALTinterwordstretchfactor\fontdimen3\font minus
  \fontdimen4\font\relax}
\providecommand\BIBforeignlanguage[2]{{%
\expandafter\ifx\csname l@#1\endcsname\relax
\typeout{** WARNING: IEEEtran.bst: No hyphenation pattern has been}%
\typeout{** loaded for the language `#1'. Using the pattern for}%
\typeout{** the default language instead.}%
\else
\language=\csname l@#1\endcsname
\fi
#2}}

\bibitem{application4}
R.~Zah{\'\i}nos, H.~Abaunza, J.~Murillo, M.~Trujillo, and A.~Viguria,
  ``Cooperative multi-uav system for surveillance and search\&rescue operations
  over a mobile 5g node,'' in \emph{International Conference on Unmanned
  Aircraft Systems (ICUAS)}.\hskip 1em plus 0.5em minus 0.4em\relax IEEE, 2022,
  pp. 1016--1024.

\bibitem{application4-1}
M.~Petrl{\'\i}k, V.~Von{\'a}sek, and M.~Saska, ``Coverage optimization in the
  cooperative surveillance task using multiple micro aerial vehicles,'' in
  \emph{International Conference on Systems, Man and Cybernetics (SMC)}.\hskip
  1em plus 0.5em minus 0.4em\relax IEEE, 2019, pp. 4373--4380.

\bibitem{application4-4}
K.~Peng, J.~Du, F.~Lu, Q.~Sun, Y.~Dong, P.~Zhou, and M.~Hu, ``A hybrid genetic
  algorithm on routing and scheduling for vehicle-assisted multi-drone parcel
  delivery,'' \emph{IEEE Access}, vol.~7, pp. 49\,191--49\,200, 2019.

\bibitem{reciprocalvo}
J.~Van~den Berg, M.~Lin, and D.~Manocha, ``Reciprocal velocity obstacles for
  real-time multi-agent navigation,'' in \emph{International conference on
  robotics and automation}.\hskip 1em plus 0.5em minus 0.4em\relax Ieee, 2008,
  pp. 1928--1935.

\bibitem{b11}
S.~Huang and K.~Low, ``A path planning algorithm for smooth trajectories of
  unmanned aerial vehicles via potential fields,'' in \emph{International
  Conference on Control, Automation, Robotics and Vision (ICARCV)}.\hskip 1em
  plus 0.5em minus 0.4em\relax IEEE, 2018, pp. 1677--1684.

\bibitem{pso1}
V.~Hoang, M.~D. Phung, T.~H. Dinh, and Q.~P. Ha, ``Angle-encoded swarm
  optimization for uav formation path planning,'' in \emph{International
  Conference on Intelligent Robots and Systems (IROS)}.\hskip 1em plus 0.5em
  minus 0.4em\relax IEEE, 2018, pp. 5239--5244.

\bibitem{MADDPG_oda1}
Y.~Zhang, Z.~Wu, Y.~Ma, R.~Sun, and Z.~Xu, ``Research on autonomous formation
  of multi-uav based on maddpg algorithm,'' in \emph{International Conference
  on Control \& Automation (ICCA)}.\hskip 1em plus 0.5em minus 0.4em\relax
  IEEE, 2022, pp. 249--254.

\bibitem{COMA}
J.~Foerster, G.~Farquhar, T.~Afouras, N.~Nardelli, and S.~Whiteson,
  ``Counterfactual multi-agent policy gradients,'' in \emph{Proceedings of the
  AAAI Conference on Artificial Intelligence}, vol.~32, no.~1, 2018.

\bibitem{Shapley}
\BIBentryALTinterwordspacing
J.~Wang, Y.~Zhang, T.-K. Kim, and Y.~Gu, ``Shapley q-value: A local reward
  approach to solve global reward games,'' \emph{Proceedings of the AAAI
  Conference on Artificial Intelligence}, vol.~34, no.~05, p. 7285–7292, Apr
  2020. [Online]. Available: \url{http://dx.doi.org/10.1609/aaai.v34i05.6220}
\BIBentrySTDinterwordspacing

\bibitem{VDN}
P.~Sunehag, G.~Lever, A.~Gruslys, W.~M. Czarnecki, V.~Zambaldi, M.~Jaderberg,
  M.~Lanctot, N.~Sonnerat, J.~Z. Leibo, K.~Tuyls, \emph{et~al.},
  ``Value-decomposition networks for cooperative multi-agent learning,''
  \emph{arXiv preprint arXiv:1706.05296}, 2017.

\bibitem{Qmix}
T.~Rashid, M.~Samvelyan, C.~S. de~Witt, G.~Farquhar, J.~Foerster, and
  S.~Whiteson, ``Qmix: Monotonic value function factorisation for deep
  multi-agent reinforcement learning,'' 2018.

\bibitem{divergence}
J.~Wang, Z.~Ren, B.~Han, J.~Ye, and C.~Zhang, ``Towards understanding linear
  value decomposition in cooperative multi-agent q-learning,'' 2020.

\bibitem{Qtran}
K.~Son, D.~Kim, W.~J. Kang, D.~E. Hostallero, and Y.~Yi, ``Qtran: Learning to
  factorize with transformation for cooperative multi-agent reinforcement
  learning,'' in \emph{International conference on machine learning}.\hskip 1em
  plus 0.5em minus 0.4em\relax PMLR, 2019, pp. 5887--5896.

\bibitem{Qplex}
J.~Wang, Z.~Ren, T.~Liu, Y.~Yu, and C.~Zhang, ``Qplex: Duplex dueling
  multi-agent q-learning,'' \emph{arXiv preprint arXiv:2008.01062}, 2020.

\bibitem{hyb5}
L.~Parker, J.~Butterworth, and S.~Luo, ``Fly safe: Aerial swarm robotics using
  force field particle swarm optimisation,'' \emph{arXiv preprint
  arXiv:1907.07647}, 2019.

\bibitem{geneticapplication1}
R.~L. Galvez, E.~P. Dadios, and A.~A. Bandala, ``Path planning for quadrotor
  uav using genetic algorithm,'' in \emph{International Conference on Humanoid,
  Nanotechnology, Information Technology, Communication and Control,
  Environment and Management (HNICEM)}.\hskip 1em plus 0.5em minus 0.4em\relax
  IEEE, 2014, pp. 1--6.

\bibitem{pathfinding1}
P.~E. Hart, N.~J. Nilsson, and B.~Raphael, ``A formal basis for the heuristic
  determination of minimum cost paths,'' \emph{Transactions on Systems Science
  and Cybernetics}, vol.~4, no.~2, pp. 100--107, 1968.

\bibitem{pathfinding2}
A.~Stentz, ``Optimal and efficient path planning for partially known
  environments,'' in \emph{Intelligent unmanned ground vehicles}.\hskip 1em
  plus 0.5em minus 0.4em\relax Springer, 1997, pp. 203--220.

\bibitem{pathfinding3}
S.~Koenig and M.~Likhachev, ``Fast replanning for navigation in unknown
  terrain,'' \emph{Transactions on Robotics}, vol.~21, no.~3, pp. 354--363,
  2005.

\bibitem{iql_oda1}
G.~Raja, S.~Anbalagan, V.~S. Narayanan, S.~Jayaram, and
  A.~Ganapathisubramaniyan, ``Inter-uav collision avoidance using
  deep-q-learning in flocking environment,'' in \emph{Ubiquitous Computing,
  Electronics \& Mobile Communication Conference (UEMCON)}.\hskip 1em plus
  0.5em minus 0.4em\relax IEEE, 2019, pp. 1089--1095.

\bibitem{vdn_oda1}
A.~Viseras, M.~Meissner, and J.~Marchal, ``Wildfire front monitoring with
  multiple uavs using deep q-learning,'' \emph{IEEE Access}, 2021.

\bibitem{qmix_oda1}
X.~Wang, M.~Yi, J.~Liu, Y.~Zhang, M.~Wang, and B.~Bai, ``Cooperative data
  collection with multiple uavs for information freshness in the internet of
  things,'' \emph{Transactions on Communications}, 2023.

\bibitem{shapley_oda1}
Z.~Liu, Z.~Wu, and Z.~Zheng, ``A cooperative game approach for assessing the
  collision risk in multi-vessel encountering,'' \emph{Ocean Engineering}, vol.
  187, p. 106175, 2019.

\bibitem{coma_oda1}
A.~Mete, M.~Mouhoub, and A.~M. Farid, ``Coordinated multi-robot exploration
  using reinforcement learning,'' in \emph{International Conference on Unmanned
  Aircraft Systems (ICUAS)}.\hskip 1em plus 0.5em minus 0.4em\relax IEEE, 2023,
  pp. 265--272.

\bibitem{huang2021}
S.~Huang, H.~Zhang, and Z.~Huang, ``E2coop: Energy efficient and cooperative
  obstacle detection and avoidance for uav swarms,'' in \emph{Proceedings of
  the International Conference on Automated Planning and Scheduling}, vol.~31,
  2021, pp. 634--642.

\bibitem{suttonpg}
R.~S. Sutton, D.~A. McAllester, S.~P. Singh, and Y.~Mansour, ``Policy gradient
  methods for reinforcement learning with function approximation,'' in
  \emph{Advances in neural information processing systems}, 2000, pp.
  1057--1063.

\bibitem{dpg}
D.~Silver, G.~Lever, N.~Heess, T.~Degris, D.~Wierstra, and M.~Riedmiller,
  ``Deterministic policy gradient algorithms,'' in \emph{International
  conference on machine learning}.\hskip 1em plus 0.5em minus 0.4em\relax PMLR,
  2014, pp. 387--395.

\end{thebibliography}

\end{document}